\documentclass[runningheads]{llncs}
\usepackage{amsmath}
\usepackage{enumitem}
\usepackage{hyphenat}
\usepackage{hyperref}

\usepackage{caption}
\usepackage{subcaption}

\mathchardef\mhyphen="2D

\usepackage{graphicx}
\usepackage[ruled,vlined]{algorithm2e}
\begin{document}

\title{WIDAR - Weighted Input Document Augmented ROUGE}
%
%
\author{Raghav Jain\inst{1}\thanks{means equal contribution.} \and
Vaibhav Mavi\inst{2}$^\star$ \and
Anubhav Jangra\inst{1}$^\star$ \and Sriparna Saha\inst{1}}
%
\institute{Indian Institute of Technology Patna, India \and
New York University, United States \\
\email{\{raghavjain106, vaibhavg152, anubhav0603, sriparna.saha\}@gmail.com}}
%
\maketitle              

\vspace{-2em}
\begin{abstract}


The task of automatic text summarization has gained a lot of traction due to the recent advancements in machine learning techniques. However, evaluating the quality of a generated summary remains to be an open problem. The  literature has widely adopted Recall-Oriented Understudy for Gisting Evaluation (ROUGE) as the standard evaluation metric for summarization. However, ROUGE has some long-established limitations; a major one being its dependence on the availability of good quality reference summary. In this work, we propose the metric WIDAR which in addition to utilizing the reference summary uses also the input document in order to evaluate the quality of the generated summary. The proposed metric is versatile, since it is designed to adapt the evaluation score according to the quality of the reference summary. The proposed metric correlates better than ROUGE by 26\%, 76\%, 82\%, and 15\%, respectively, in coherence, consistency, fluency, and relevance on human judgement scores provided in the SummEval dataset. The proposed metric is able to obtain comparable results with other state-of-the-art metrics while requiring a relatively short computational time\footnote{Implementation for WIDAR can be found at - \href{https://github.com/Raghav10j/WIDAR}{https://github.com/ Raghav10j/WIDAR}}.
\vspace{-0.7em}
\vspace{-1em}
\end{abstract}

\section{Introduction} \label{sec:intro}
\vspace{-0.7em}
Accessibility of internet has led to massive increase in content available to a user, making it difficult to obtain the required information. This seemingly perpetual growth of information necessitates the need for automatic text summarization tools. Text summarization can be described as the task of generating fluent and human readable summary while preserving the essence of the original text documents. Evaluation of these automatically generated summaries has been actively explored by the research community for over 5 decades \cite{Edmundson1969NewMI}. Since then, various attempts have been made to quantify the effectivenes of the summarization systems; however the evaluation task still remains an open problem till this day. \par

\begin{figure}
    \centering
    \includegraphics[width=\linewidth, scale=0.3]{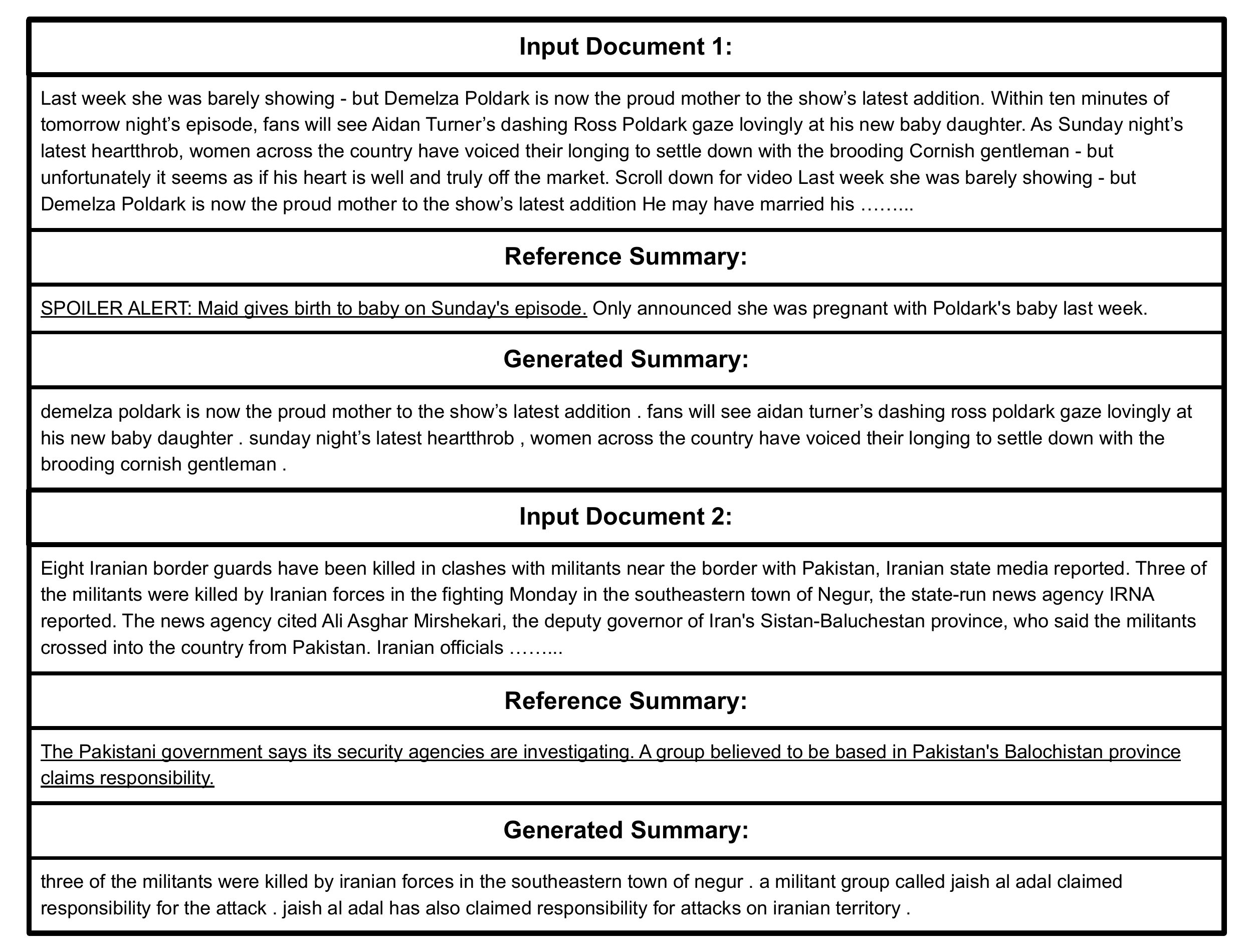}
    \vspace{-6mm}
    \caption{Examples from DailyMail/CNN dataset \cite{Nallapati2016AbstractiveTS} where ground truth is unsatisfactory either due to clickbaits (Eg.-1), or information incompleteness (Eg.-2).}
    \label{fig:example}
\end{figure}

The most widely adopted evaluation metric for text summarization in the community is \textit{Recall-Oriented Understudy for Gisting Evaluation (ROUGE)} \cite{lin-2004-rouge} which is mainly based on the n-gram overlap between the generated summary and reference summary. However, ROUGE's dependency on a good quality reference summary is one of it's biggest drawback. Fabbri et al. \cite{fabbri2021summeval} highlighted the inconsistency in quality of some reference summaries in the CNN/DailyMail dataset  \cite{Nallapati2016AbstractiveTS} by describing the summaries consisting of clickbaits instead of being truthful and informative with respect to the input article (refer to Fig. \ref{fig:example}). Kryscinski et al. \cite{kryscinski-etal-2019-neural} also reported this issue of reference summaries containing irrelevant information such as links to other articles or factual inconsistency in the Newsroom dataset \cite{grusky-etal-2018-newsroom}. Even if a reference summary is of satisfactory quality, it is highly unlikely that it is the only acceptable summary of that document as different people tend to produce different summaries for the same document \cite{Nenkova2006SummarizationEF,Rath1961TheFO}. Therefore, all the above-mentioned claims imply that sole dependence on reference summary for an evaluation metric is not optimal. Therefore, we propose an evaluation metric that also considers an input source document while evaluating the quality of its summary. \par 

\vspace{-2em}
\vspace{-2em}

In order to design an evaluation metric, it is important to study what comprises of a good summary. Ideally, a summary must be \textit{coherent}, \textit{non-redundant}, \textit{fluent}, \textit{consistent} and \textit{relevant} to the input article \cite{dang2005overview}. Using these characteristics, recent works have attempted to quantify and compare the performance of existing evaluation metrics  \cite{Bhandari-2020-reevaluating,fabbri2021summeval}.  These works highlight the limitations of existing metrics and offer various resources for conducting further research on the evaluation task. One such work is the SummEval dataset  \cite{fabbri2021summeval} that provides human annotation scores for - \textit{coherence}, \textit{consistency}, \textit{fluency} and \textit{relevance}. \par

\begin{figure}
    \centering
    \includegraphics[width=0.7\linewidth, scale=0.3]{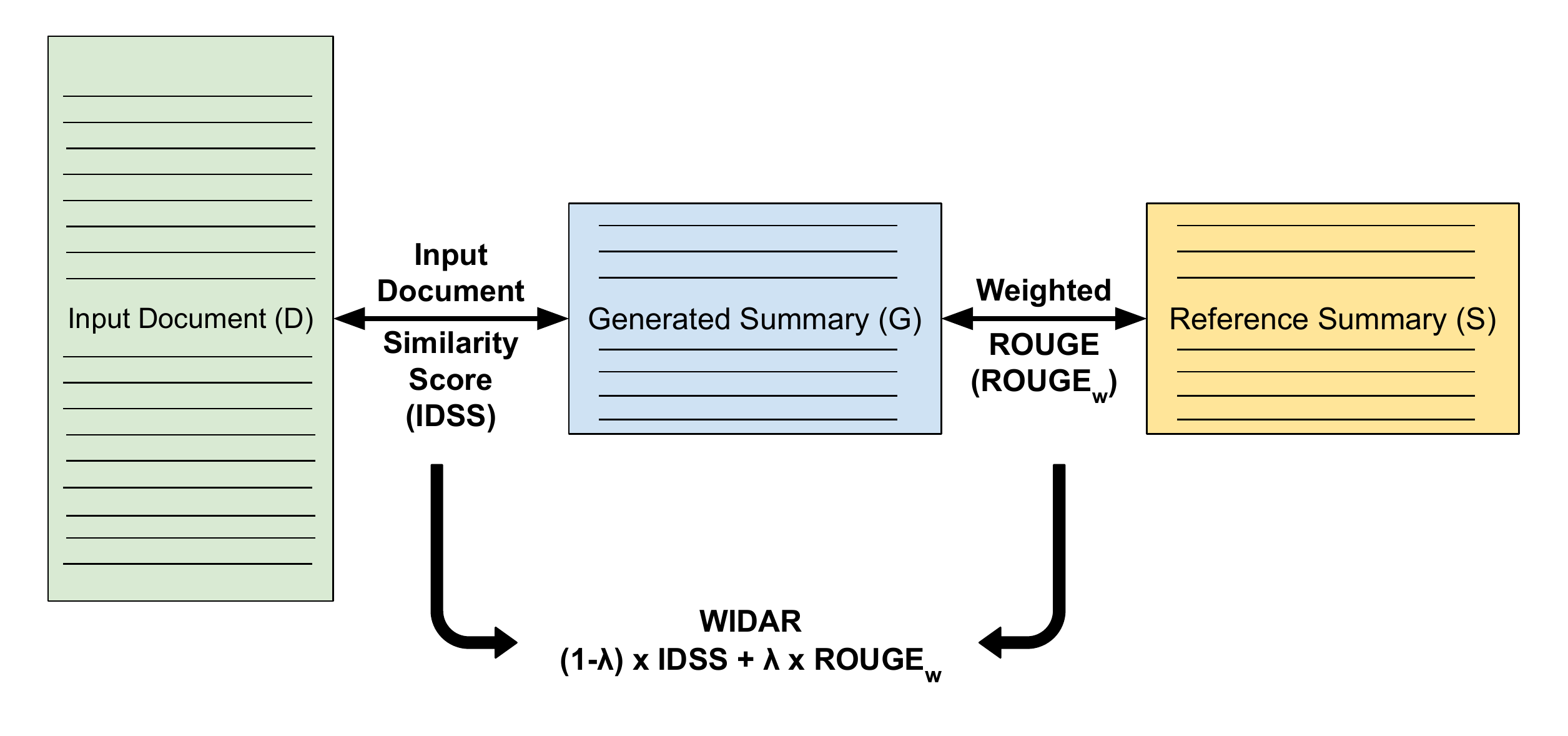}
    \vspace{-2mm}
    \caption{Model figure for WIDAR.}
    \label{fig:box}
\end{figure}

In this paper, we propose an evaluation metric \textit{WIDAR (Weighted Input Document Augmented ROUGE)} in an attempt to overcome the above-mentioned limitations of ROUGE (refer to Fig. \ref{fig:box}). The proposed metric utilizes the reference summary and input document to measure the quality of a generated summary. WIDAR introduces the idea of weighted ROUGE that relies on weighting sentences in reference summary based on information coverage and redundancy within the summary. Through experiments, we illustrate that WIDAR is able to outperform ROUGE by a large margin, and is able to obtain comparable results with other state-of-the-art metrics while requiring relatively short computational time. \par

\vspace{-1.3em}

\vspace{-4em}
\section{Related Works} \label{sec:related_works}
\vspace{-0.7em}
The approaches to text summarization can be broadly classified into two categories, extractive methods \cite{10.1145/215206.215333,saini2019extractive,Paice1990ConstructingLA} and abstractive methods \cite{chopra-etal-2016-abstractive,46111,jangra2020semantic}. Summarization research today has expanded into more complex problems like multi-lingual summarization \cite{hasan2021xl,scialom2020mlsum}, multi-modal summarization \cite{jangra2020text,jangra2021survey,jangra2020multimodal,jangra2021multi}, across-time summarization \cite{duan2019across} etc.


Numerous evaluation metrics have been proposed to assess summarization systems. Some of them are based on text matching between predicted summary and reference summary such as Recall-Oriented Understudy for Gisting Evaluation (ROUGE) \cite{lin-2004-rouge}, ParaEval \cite{zhou-etal-2006-paraeval}, ROUGE 2.0 \cite{Ganesan2018ROUGE2U}, Metric for Evaluation of Translation with Explicit ORdering (METEOR) \cite{lavie-agarwal-2007-meteor}, Bilingual Evaluation Understudy (BLEU) score \cite{Papineni2002BleuAM},  Character n-gram F-score (CHRF) \cite{popovic-2015-chrf}, Consensus-based Image Description Evaluation (CIDEr) \cite{Vedantam2015CIDErCI} etc. There are also evaluation metrics that try to capture semantic similarity including word embeddings based techniques such as Word Mover similarity (WMS)  \cite{pmlr-v37-kusnerb15}, MoverScore \cite{zhao-etal-2019-moverscore}, Sentence Mover Similarity (SMS)  \cite{clark-etal-2019-sentence}, ROUGE-WE \cite{ng-abrecht-2015-better}, ELMo-m \cite{sun-nenkova-2019-feasibility}, automated pyramid metric \cite{passonneau-etal-2013-automated} and graph based techniques such as graph based ROUGE (ROUGE-G) \cite{shafieibavani-etal-2018-graph} and AUTOmatic SUMMary Evaluation based
on N-gram Graphs (AutoSummENG) \cite{Giannakopoulos2011AutoSummENGAM}. Other than these, there are also model based learned metrics such as Supervised Summarization Scorer (S$^3$) \cite{peyrard-etal-2017-learning}, BERTScore \cite{Zhang*2020BERTScore:}, NeuralTD \cite{Bhm2019BetterRY}, Support Vector
Regression (SVR)  \cite{ShafieiBavani2018SummarizationEI} and question answering based metrics such as Answering
Performance for Evaluation of Summaries (APES) \cite{eyal-etal-2019-question} and Semantic QA \cite{Chen2018ASQ}. In unsupervised settings where evaluation is carried out on the basis of input document rather than depending on a reference summary, SummaQA \cite{scialom-etal-2019-answers}, summarization evaluation with pseudo references and BERT (SUPERT) \cite{gao2020supert} and BLANC \cite{Lita2005BLANCLE} are some of the most recent and state-of-the-art metrics.

\vspace{-1.5em}
\section{WIDAR Evaluation Metric} \label{sec:proposed_method}
\vspace{-1.2em}
We propose WIDAR (Weighted  Input Document Augmented ROUGE), an evaluation metric that utilizes both reference summary ($R$) and input document ($D$) to judge the quality of a generated summary ($S$). For better understanding, we divide our approach into two steps: 1) calculation of Weighted ROUGE (Section \ref{sec:Weighted_ROUGE}), and 2) combination of Weighted ROUGE with similarity score computed between generated summary and input document to obtain WIDAR (Section \ref{sec:combining}). Table \ref{tab:Notation} lists the notations used in the remainder of this paper. 


\vspace{-3em}
\begin{table}
  \centering
  \caption{Notation of each variable and its corresponding meaning.}
  \label{tab:Notation}
    \begin{tabular}{cl}
    \hline
    \textbf{Notation} & \textbf{Meaning}      \\
    \hline
    $D$ & input document \\
    $R$ & reference summary \\
    $S$ & generated summary\\
    \hline
    
    $d_i$ & $i^{th}$ input document's sentence \\
    $r_i$ & $i^{th}$ input reference summary's sentence \\
    $s_i$ & $i^{th}$ input generated summary's sentence \\
    \hline
    
    $w_{cov_i}$ & coverage weight assigned to $i^{th}$ generated summary sentence \\
    $w_{red_i}$ & redundancy weight assigned to $i^{th}$ generated summary sentence \\
    $w_i$ & overall weight assigned to $i^{th}$ generated summary sentence \\
    \hline
    \vspace{-3.5em}
    \end{tabular}%
\end{table}
\vspace{-2em}
\subsection{Weighted ROUGE} \label{sec:Weighted_ROUGE}
\vspace{-1em}
As discussed in Section \ref{sec:intro}, ROUGE is highly dependent on the quality of reference summary to perform effectively. However, in real world scenarios, high quality of reference summary is not assured.. Therefore, we introduce two special weights for each reference summary sentence to penalize/reward the quality of information present in this sentence. Each reference summary sentence $r_i$ is assigned two scores: 1) Coverage weight ($w_{cov_i}$) - based on the input document information that is covered by $r_i$, and 2) Redundancy weight ($w_{red_i}$) - based on the uniqueness of information presented by $r_i$ in the reference summary. We use Algorithm \ref{algo:Ws}\footnote{Here $\theta_1$ and $\theta_2$ are ROUGE-L thresholds for coverage and redundancy respectively.} to compute the redundancy weights and coverage weights for all sentences in the reference summary. We then obtain the overall weight\footnote{We multiply the final weights by the number of sentences in the reference summary $|R|$ to ensure that the sum of weights remains the same as in plain ROUGE, i.e., $\sum_i w_i = |R|$.} ($w_i$) for $r_i$ by computing the average of $w_{cov_i}$ and $w_{red_i}$.

\begin{equation} \label{eq:FW}
    w_{i} = \dfrac{(w_{cov_{i}}+w_{red_{i}})} {2} \times |R|
\end{equation}

\begin{algorithm}[H]
\caption{\small{Calculating the coverage and redundancy weights.}}
\label{algo:Ws}
\SetAlgoLined
\KwIn{$R = \{r_i\}$, $D = \{d_j\}$}
\KwOut{$W_{cov} = \{w_{cov_i}\}$ $W_{red} = \{w_{red_i}\}$}
$W_{cov}$, $W_{red}$ $\leftarrow$ emptyList\;
\For{$r_i$ in R}{
    $w_{cov_i}$ = 0\;
    \For{$d_j$ in D}{
    \If{$ROUGE\mhyphen L^{r}$($r_i$,$d_j$)$ \geq \theta_1$}{
    $w_{cov_i}$++\;
    }
    }
    $W_{cov}$ $\leftarrow$ $w_{cov_i}$/$|D|$\;
    
}
\For{$r_i$ in R}{
        $w_{red_i}$ = 0\;
        \For{$r_j$ in R}{
            \If{$r_i \neq r_j$ \& $ROUGE\mhyphen L^{r}$($r_i$,$r_j)$ $ \geq \theta_2$}{
                    $w_{red_i}$++\;
                }
        }
        $W_{red}$ $\leftarrow$ 1-($w_{red_i}$ / $|R|$)\;
}
\end{algorithm}

We propose sentence-level ROUGE-N ($ROUGE\mhyphen N_{SL}$) and sentence-level ROUGE-L ($ROUGE\mhyphen L_{SL}$), variations of ROUGE in order to incorporate the sentence-level redundancy and coverage weights (Eq. \ref{eq:FW}), respectively. \\
\noindent\underline{Sentence-level ROUGE-N:} Typically, ROUGE-N measures the number of overlapping n-grams between the reference summary and the generated summary. However, to compute the sentence-level ROUGE-N ($ROUGE\mhyphen N_{SL}$) we take into account sentence level n-grams for the overlap count, \textit{viz.} we discard the bridge n-grams (that share words from two or more sentences)\footnote{Note that $ROUGE\mhyphen 1$ and $ROUGE\mhyphen 1_{SL}$ denote the same metrics.}. We use the following equations to measure the precision ($ROUGE\mhyphen N_{SL}^{p}$), recall ($ROUGE\mhyphen N_{SL}^{r}$), and f-score ($ROUGE\mhyphen N_{SL}^{f}$), respectively.

\begin{equation}\label{eq:SRR}
    ROUGE\mhyphen N_{SL}^{r} =  \dfrac{\sum_{s\mhyphen gram_i}\sum_{r\mhyphen gram_j}count(s\mhyphen gram_i,r\mhyphen gram_j) }{\sum_{r\mhyphen gram_j} |r\mhyphen gram_j|}
\end{equation}
\begin{equation}\label{eq:SRP}
    ROUGE\mhyphen N_{SL}^{p} =  \dfrac{\sum_{s\mhyphen gram_i}\sum_{r\mhyphen gram_j}count(s\mhyphen gram_i,r\mhyphen gram_j) }{\sum_{s\mhyphen gram_j} |s\mhyphen gram_j|}
\end{equation}
\begin{equation}\label{eq:SRF}
       ROUGE\mhyphen N_{SL}^{f}=
       \dfrac{2 \times (ROUGE\mhyphen N_{SL}^{r}) \times (ROUGE\mhyphen N_{SL}^{p})}   {(ROUGE\mhyphen N_{SL}^{r})+(ROUGE\mhyphen N_{SL}^{p})}
\end{equation}
where $s\mhyphen gram_i$ and $r\mhyphen gram_i$ denote the sentence-level n-grams for $i^{th}$ sentence in the generated summary and in the reference summary, respectively; $count(s\mhyphen gram_i,$ $r\mhyphen gram_j)$ calculates the number of overlapping n-grams in $s\mhyphen gram_i$ and $r\mhyphen gram_i$, and $|.|$ denotes the cardinality of a set.

\noindent\underline{Sentence-level ROUGE-L:} ROUGE-L computes the longest common sub-sequence of words between the generated summary and the reference summary. Sentence-level ROUGE-L ($ROUGE\mhyphen L_{SL}$) is computed as follows:

\begin{equation}\label{eq:SRLR}
    ROUGE\mhyphen L_{SL}^{r} =  \dfrac{\sum_{r_i\in R} UnionLCS(r_i, S)}{|R|}
\end{equation}

\begin{equation}\label{eq:SRLP}
    ROUGE\mhyphen L_{SL}^{p} =  \dfrac{\sum_{r_i\in R} UnionLCS(r_i, S)}{|S|}
\end{equation}

\begin{equation}\label{eq:SRLF}
       ROUGE\mhyphen L_{SL}^{f}=
       \dfrac{2 \times (ROUGE\mhyphen L_{SL}^{r}) \times (ROUGE\mhyphen L_{SL}^{p})}   {(ROUGE\mhyphen L_{SL}^{r})+(ROUGE\mhyphen L_{SL}^{p})}
\end{equation}
where $UnionLCS(r_i, S)$ is the union of the longest common sub-sequence computed between a reference summary sentence ($r_i\in R$) and each sentence of generated summary ($s_i\in S$), and $|R|$ and $|S|$ denote the number of sentences in reference summary and generated summary, respectively. \par

We integrate into these sentence-level ROUGE metrics the weights (Eq. \ref{eq:FW}) to obtain Weighted ROUGE-N ($ROUGE\mhyphen N_{W}$) and Weighted ROUGE-L ($ROUGE\mhyphen L_W$) scores. $ROUGE\mhyphen N_W$ is obtained by multiplying $w_i$ in each summation term in Eqs. \ref{eq:SRR} to \ref{eq:SRF}, and $ROUGE\mhyphen L_W$ is obtained by multiplying $w_i$ in each summation term in Eqs. \ref{eq:SRLR} to \ref{eq:SRLF}. 

\vspace{-0.7em}
\subsection{Combining Weighted ROUGE with Input Document Similarity}
\vspace{-0.5em}
\label{sec:combining}
\underline{Input Document Similarity Score (IDSS)} We incorporate information overlap of generated summary with input document to make the proposed metric more robust and applicable to the real-world situations where the quality of reference summary might be sometimes inadequate. For simplicity, we use ROUGE-L F-score as the similarity measure, because it performed better than other ROUGE variants in our experiments (refer to Section \ref{sec:exp_set}). Therefore,
\begin{equation}\label{eq:IDSS}
    IDSS = ROUGE\mhyphen L^f(S, D)
\end{equation}


The last step of the evaluation process is to combine the $ROUGE_W$ and $IDSS$ scores in such a way that the final score retains the individual characteristic of both the individual scores. We define $WIDAR$ as follow:

\begin{equation}\label{eq:FS1}
    WIDAR^{X}_{K} = (1-\lambda) \times IDSS + \lambda \times ROUGE\mhyphen K_{W}^{X}
\end{equation}
where x $\in$ \{r, p, f\} and K $\in \{1, 2, L\}$; $\lambda$ is a hyper-parameter directly proportional to the quality of coverage in reference summary\footnote{$\lambda$ is a fixed hyper-parameter, which is set to 0.5 in our final experiments. We attempted to make $\lambda$ a data-driven parameter by setting $\lambda = max(w_{cov_i})$ or $\lambda = mean(w_{cov_i})$, but this setting was not able to outperform the fixed $\lambda=0.5$ value (refer to Section \ref{sec:exp_set}).}.

\section{Experiments} \label{sec:experiments}
\vspace{-0.7em}
\subsection{Dataset} \label{sec:dataset}
\vspace{-0.5em}
For all the experiments conducted in this work, we have used the SummEval dataset \cite{fabbri2021summeval}. It contains the summaries generated by 23 recent summarization models trained on CNN/DailyMail dataset \cite{Nallapati2016AbstractiveTS}. The dataset contains human annotation scores for 16 generated summaries of 100 source news articles giving us 1600 summary-text pairs. Each summary is annotated by 3 experts and 5 crowd-source annotators to evaluate the quality of a summary on a range of 1-5 across 4 different characteristics: 1) \textit{Coherence:} measures the quality of smooth transition between different summary sentences such that sentences are not completely unrelated or completely same, 2) \textit{Consistency:} measures the factual correctness of summary with respect to input document, 3) \textit{Fluency:} measures the grammatical correctness and readability of sentences, 4) \textit{Relevance:} measures the ability of a summary to capture important and relevant information from the input document. Apart from human annotation scores, 11 reference summaries for each example, and evaluation scores for generated summaries across different evaluation metrics are also made available in the dataset repository\footnote{\url{https://github.com/Yale-LILY/SummEval}}.

\vspace{-1em}
\subsection{Evaluation of Evaluation Metric} \label{sec:kendall_tau}
\vspace{-0.5em}
In order to measure the performance of the proposed evaluation metric, we calculate the correlation between the scores of that metric and the average annotation scores for each characteristic of each summary for 1600 summary-text examples provided in the SummEval dataset \cite{fabbri2021summeval} (described in Section \ref{sec:dataset}). We have used the average of expert annotation scores for our experiments because of the inconsistency between expert and crowd-source scores reported by Fabbri et al. \cite{fabbri2021summeval}. We use the Kendall’s tau correlation coefficient as the correlation metric in our experiments. Kendall’s tau correlation between two sequences $X=\{x_{i}\}$ and $Y=\{y_{i}\}$ is defined as follows:
\begin{equation}\label{eq:KT}
    \tau = \dfrac{C-D}{C+D}
\end{equation}
where C is the number of all those pairs that are concordant and D is the number of all those pairs that are discordant in sequences, $X$ and $Y$.

\vspace{-1em}
\subsection{Experimental Settings} \label{sec:exp_set}
\vspace{-0.5em}
In this section, we discuss various hyperparameters used in the proposed methodology, along with the tuning experiments carried out to justify them\footnote{All the hyperparameter tuning experiments were performed using $ROUGE\mhyphen L^{f}$ unless stated otherwise.}.

\begin{figure}[!h]
\centering
\vspace{-5mm}
\begin{subfigure}{0.42\textwidth}
  \centering
  \includegraphics[width=\linewidth]{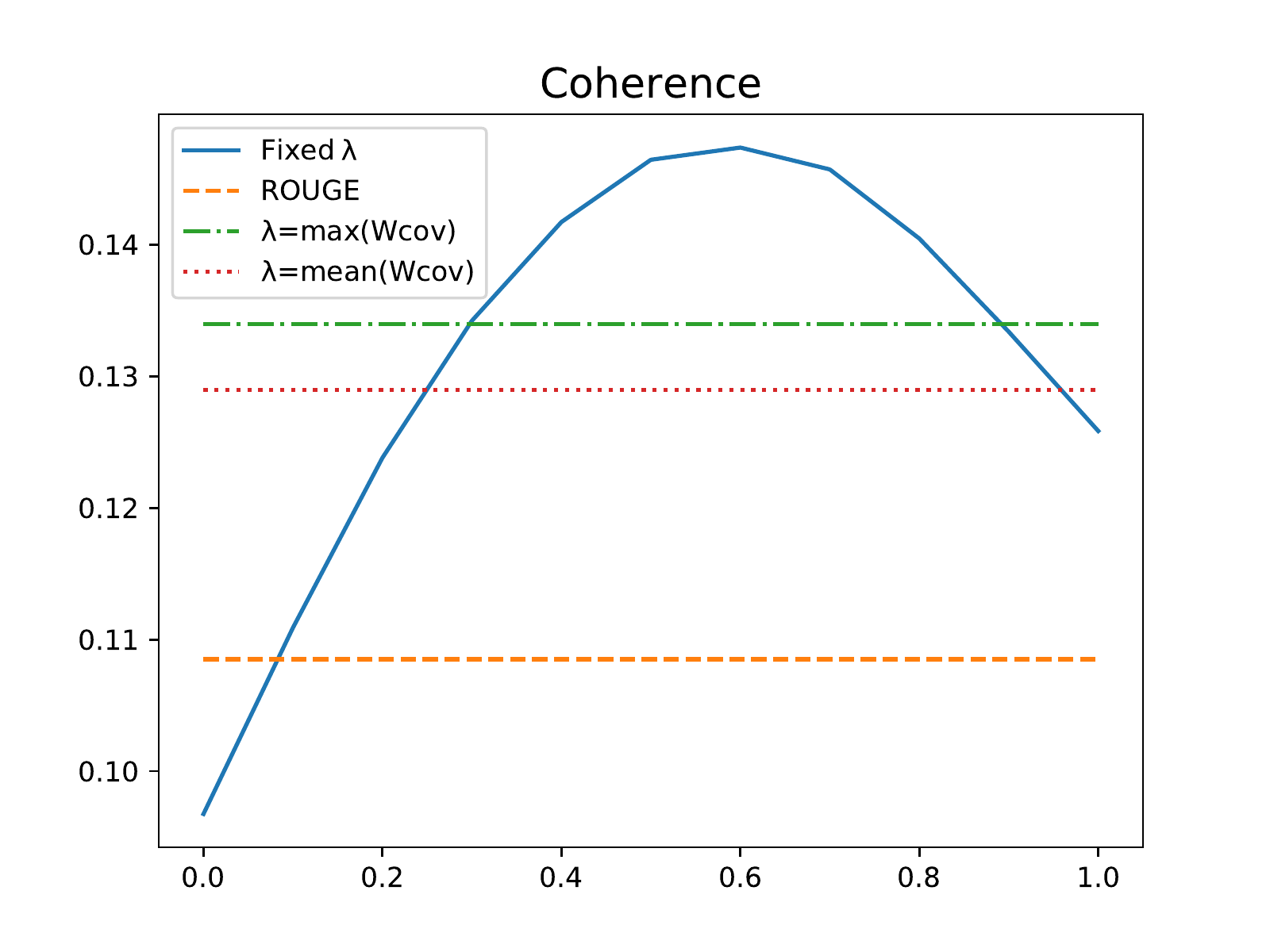}
\end{subfigure}%
\begin{subfigure}{0.42\textwidth}
  \centering
  \includegraphics[width=\linewidth]{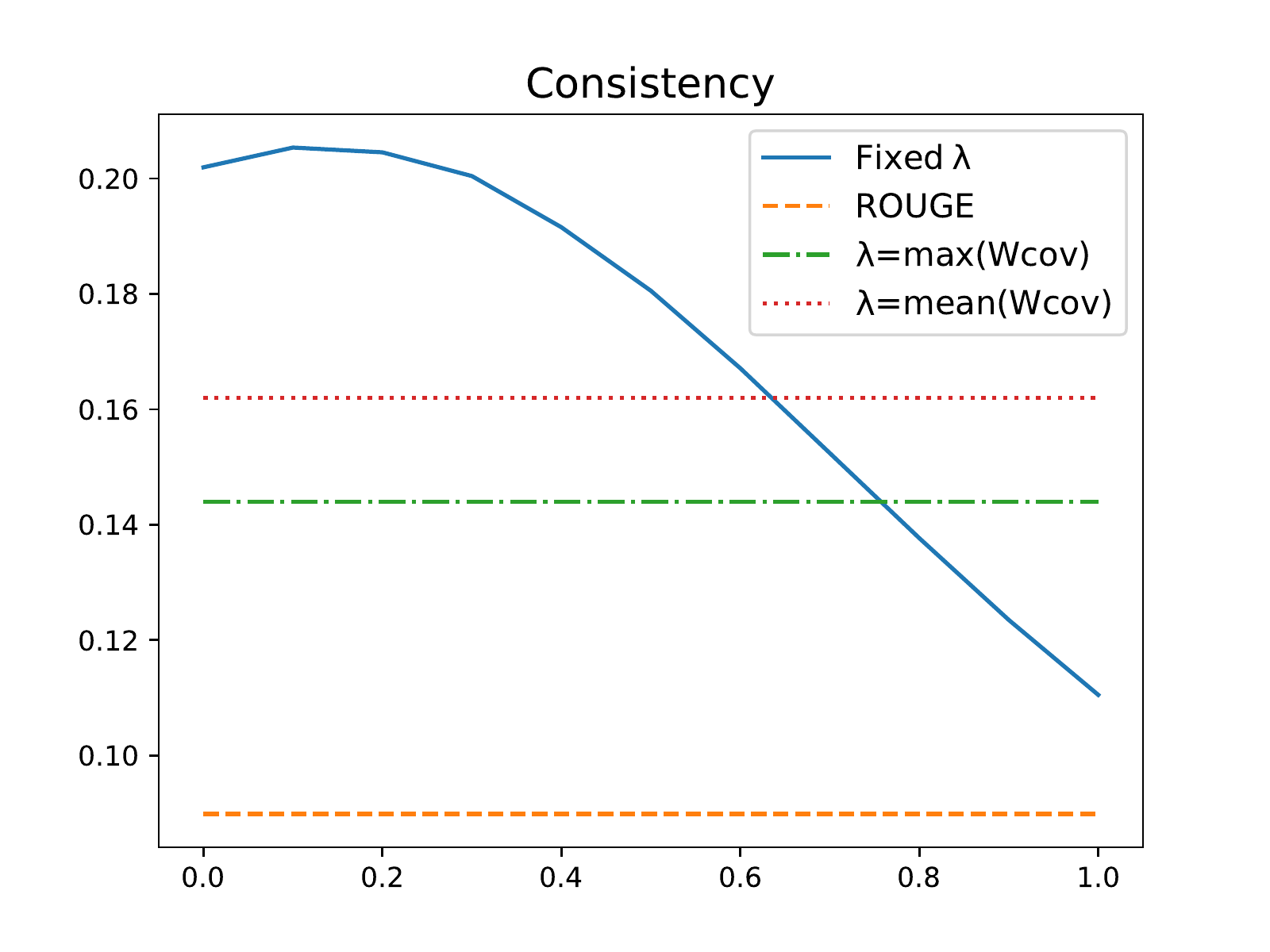}
\end{subfigure}
\newline
\begin{subfigure}{0.42\textwidth}
  \centering
  \includegraphics[width=\linewidth]{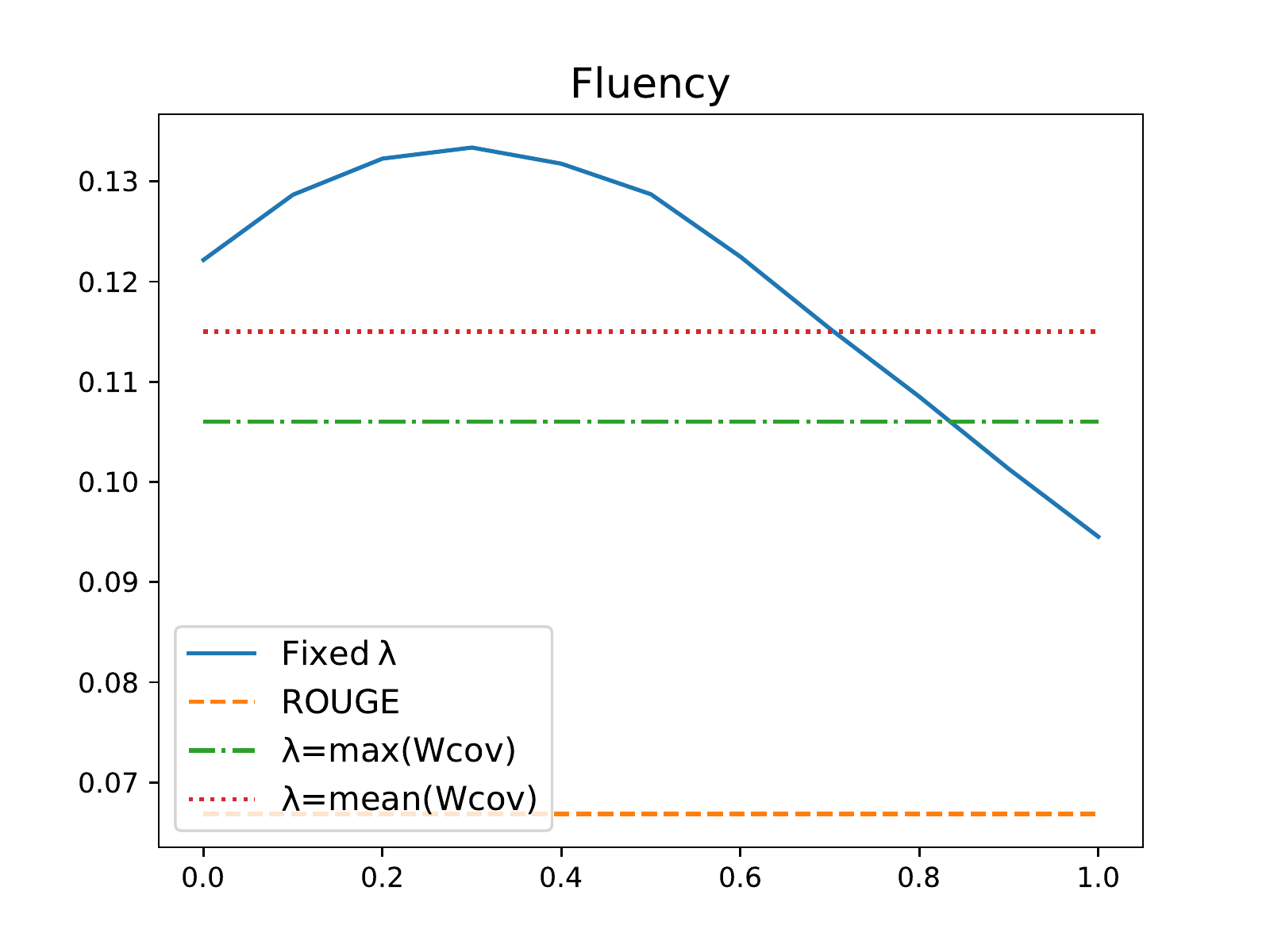}
\end{subfigure}
\begin{subfigure}{0.42\textwidth}
    \includegraphics[width=\linewidth]{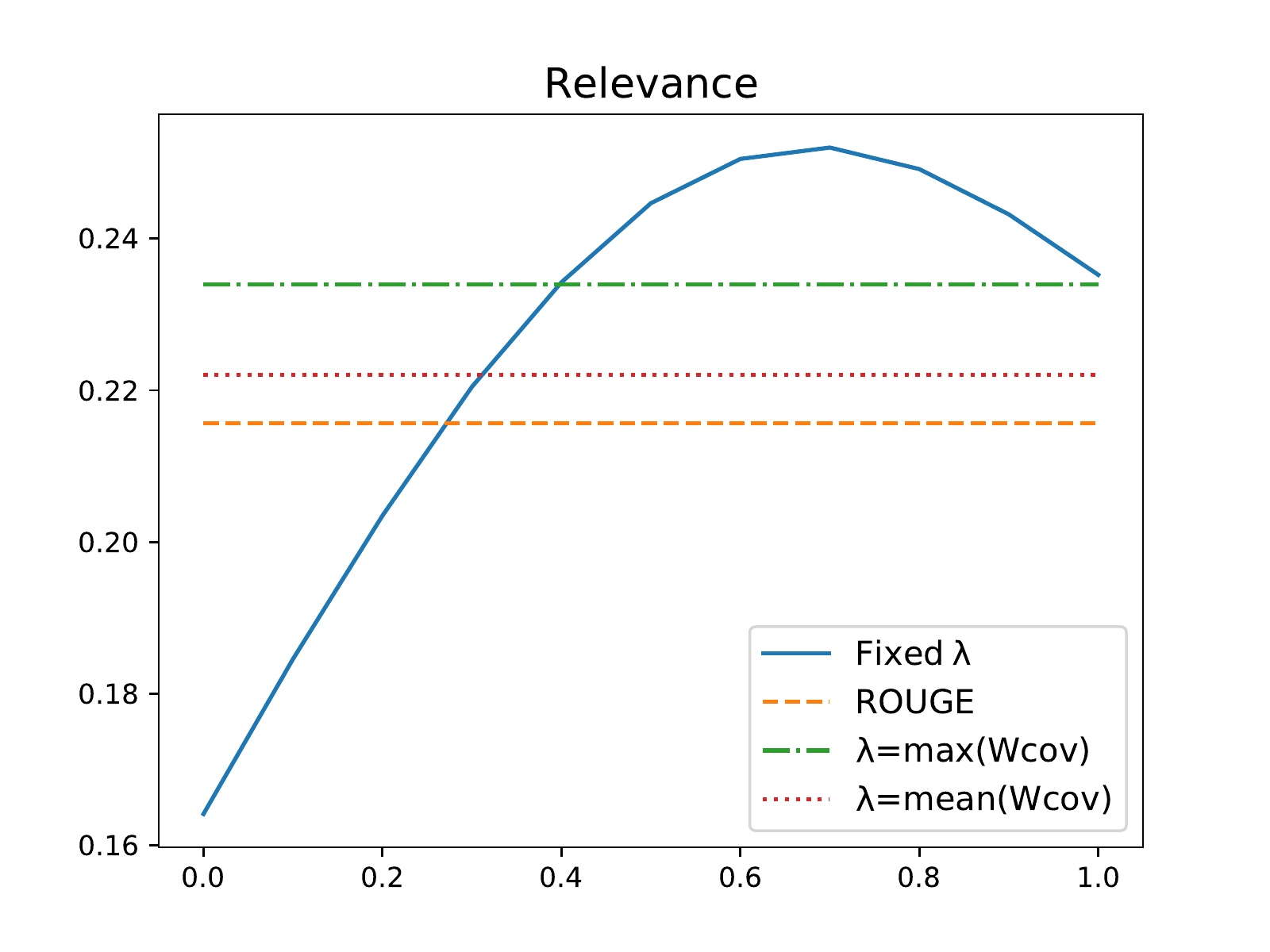}
\end{subfigure}
\vspace{-2mm}
\caption{Correlation plots of WIDAR with human judgement scores (from SummEval dataset \cite{fabbri2021summeval}) for different $\lambda$ values.}
\label{fig:lambda}
\end{figure}

\noindent\underline{Weighted sum of $IDSS$ and $ROUGE_W$ ($\lambda$):} $\lambda$ is used to get the weighted sum of information overlap of the generated summary with the input document ($IDSS$) and the reference summary ($ROUGE_W$). We attempted to investigate the optimal value of $\lambda$ using a data-driven technique. To be more precise, since $\lambda$ indicates the balance between the degree of attention given to the input document and the reference summary, we hypothesize that making $\lambda$ adapt to the information shared in reference summary and input document should give us better performance since the higher the overlap, the better the quality of summary, and the higher the $\lambda$ should be. Hence we perform two different experiments with $\lambda = max(w_{cov_i})$ and $\lambda = mean(w_{cov_i})$. To compare performance of a fixed $\lambda$ value with the defined data-driven strategy, we plot performance of the proposed technique with fixed values of $\lambda \in \{0.0, 0.1, 0.2, ... , 1.0\}$ (see Fig. \ref{fig:lambda}). Even though both of these $\lambda$ defining strategies outperform the baseline metric ROUGE, we notice that the d value of $\lambda = 0.5$ is able to outperform these data-driven strategies as well as most of the fixed $\lambda$ values\footnote{It was also noticed that $\lambda = mean(W_{cov})$ outperforms $\lambda = max(W_{cov})$ in fluency and consistency; while the opposite happens for coherence and relevance. The reason for this can be explained by the fact that $mean(W_{cov}) < max(W_{cov})$; therefore the $\lambda = mean(W_{cov})$ variation always gives more weight to the input document similarity, giving higher fluency and consistency scores because input document consists of all the informationally rich and grammatically correct sentences.}.

\vspace{-2em}

\begin{figure}[h]
\centering
\begin{subfigure}{0.45\textwidth}
  \centering
  \includegraphics[width=\linewidth]{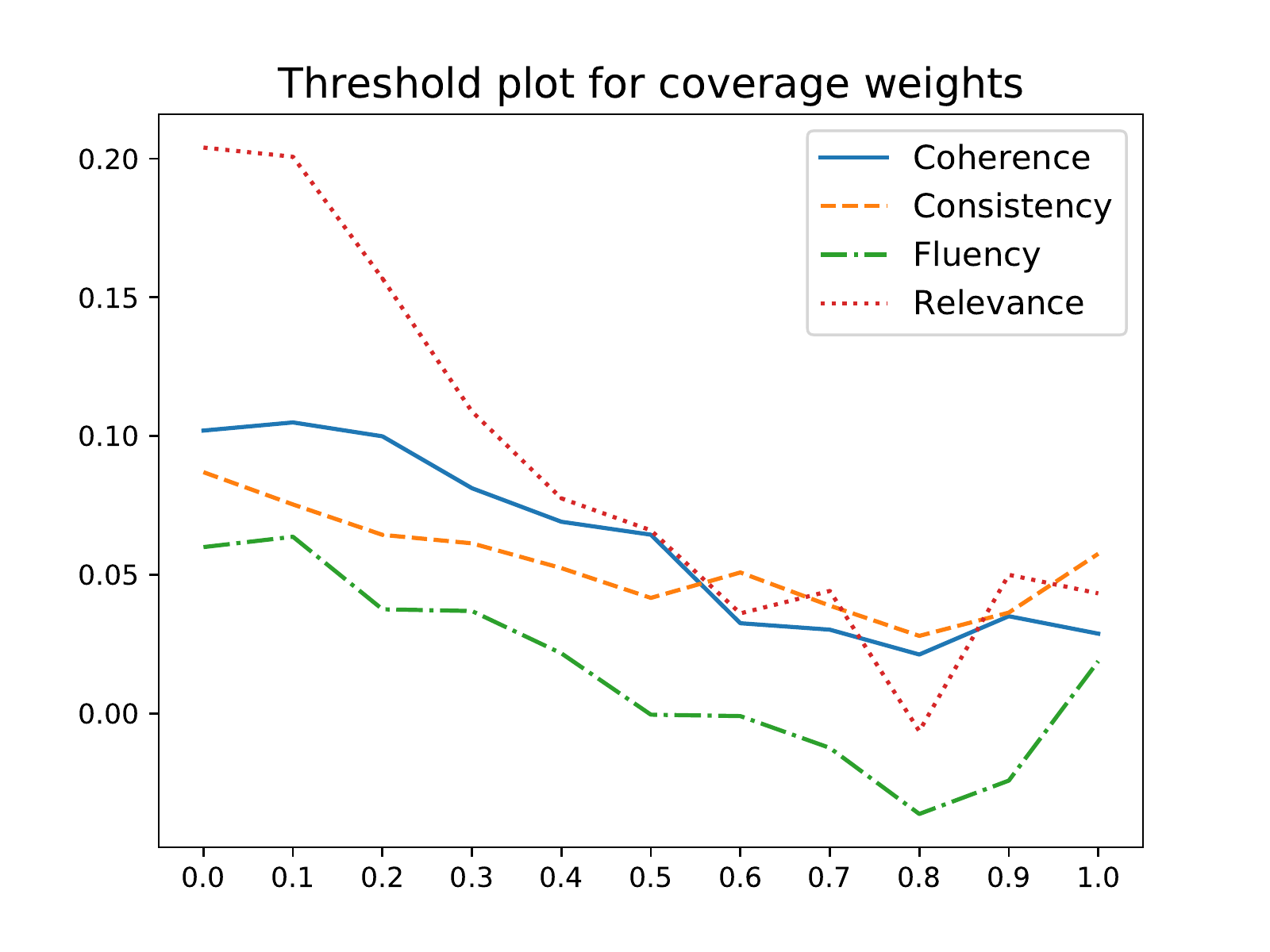}
\end{subfigure}%
\begin{subfigure}{0.45\textwidth}
  \centering
  \includegraphics[width=\linewidth]{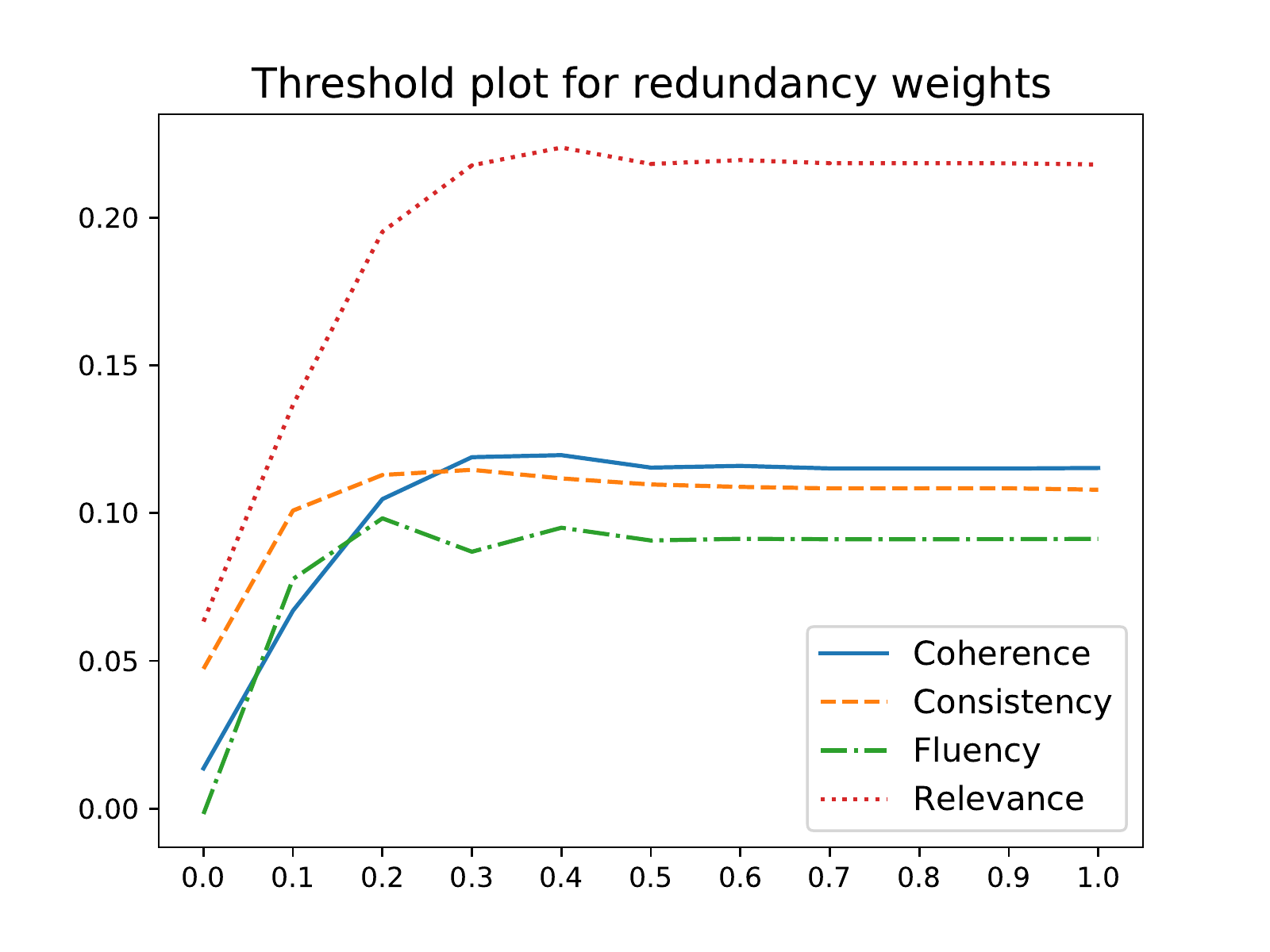}
\end{subfigure}
\vspace{-2mm}
\caption{Correlation plots of WIDAR with human judgement scores (from SummEval dataset \cite{fabbri2021summeval}) with varying $\theta_1$ values (on left) and $\theta_2$ (on right) values.}
\vspace{-2em}
\label{fig:theta}
\end{figure}

\noindent\underline{Thresholds for $w_{cov_i}$ and $w_{red_i}$:} $\theta_1$ and $\theta_2$ are the hyperparameters used in the calculation of coverage weights ($w_{cov_i}$) and redundancy weights ($w_{red_i}$) (Algorithm \ref{algo:Ws}), respectively. To obtain the optimal range of these hyperparameters; we first performed an individual search for both $\theta_1$ and $\theta_2$ (see Fig \ref{fig:theta}). As per these experiments, $\theta_1 = 0.0$ or $0.1$ and $\theta_2 = 0.4$ yielded the best results when analyzed individually. However, on further experimentation, it was found that the best performance was obtained at $\theta_1 = 0.1$ and $\theta_2 = 0.3$.

\noindent\underline{Similarity function for $IDSS$:} In order to find the most suitable similarity function to compute the information overlap between input document and generated summary, we performed an isolated experiment where correlation coefficient of  similarity function candidates was computed with the human judgement scores (Table \ref{tab:IDSS corr}). ROUGE-L$^f$ score was the best performing model, and hence chosen as the similarity function.

\vspace{-2.5em}
\begin{table}
  \centering
  \caption{Various ROUGE-based similarity functions for $IDSS$.}
  \label{tab:IDSS corr}
    \begin{tabular}{l@{\hskip 0.1in}c@{\hskip 0.1in}c@{\hskip 0.1in}c@{\hskip 0.1in}c}
    \hline
    \textbf{IDSS} & \textbf{Coherence} & \textbf{Consistency}      &\textbf{Fluency} &\textbf{Relevance} \\
    \hline

    ROUGE-1$^{r} $ & 0.033 & 0.101 & 0.050 &0.123 \\
    ROUGE-1$^{f} $ & 0.035 & 0.108 & 0.055 & 0.117 \\
    ROUGE-2$^{r} $ & 0.066 & 0.183 & 0.111 & 0.149 \\
    ROUGE-2$^{f} $ & 0.072 & 0.194 & 0.118 & 0.153 \\
    \hline
    ROUGE-L$^{r} $ & 0.088 & 0.187 & 0.112 & 0.158 \\
    ROUGE-L$^{f} $ & 0.097 & 0.202 & 0.122 & 0.164 \\

    \hline
    \end{tabular}%
\end{table}
\vspace{-3.5em}

\section{Results and Discussions} \label{sec:results}
\vspace{-1em}
We evaluate the performance of our metric with other state-of-the-art techniques using correlation coefficient described in Section \ref{sec:kendall_tau}. Table \ref{tab:WIDAR corr}\footnote{In case a metric has more than one variation, the version that corresponds to f-score was used.} lists the correlation of WIDAR and other state-of-the art-metric scores available in SummEval with human judgement scores\footnote{All the reported metrics in Table \ref{tab:WIDAR corr} have been computed in a multi-reference setting using 11 reference summaries per generated summary.}. These scores illustrate the superiority of WIDAR over its predecessor, ROUGE, by a wide margin in all the three variants. \par
It can be deduced from the results that we need a way to combine these scores to better evaluate the performance of each metric, since a metric like SMS \cite{clark-etal-2019-sentence} performs well in aspects like consistency and fluency, yet it gives mediocre performance in coherence and relevance. Therefore, we also provide the average of these four scores in an attempt to ascertain the overall performance of each metric. We find out that all three variants of WIDAR are able to perform satisfactory, as they appear as $2^{nd}$, $3^{rd}$ and $4^{th}$ in the overall rankings; as opposed to their ROUGE counter-parts that end up in the middle-bottom section of the rankings. \par

\begin{table}[t]
  \centering
  \caption{Evaluation of the proposed metric WIDAR against other state-of-the-art methods using Kendall's Tau correlation coefficient over human judgement scores of individual summary components described in SummEval dataset \cite{fabbri2021summeval}. Average denotes the average score over coherence, consistency, fluency and relevance. (.) denotes the rank of metric for the corresponding column.}
  \label{tab:WIDAR corr}
    \begin{tabular}{l@{\hskip 0.07in}c@{\hskip 0.07in}c@{\hskip 0.07in}c@{\hskip 0.07in}c@{\hskip 0.1in}c}
    \hline
    \textbf{Metric} & \textbf{Coherence} & \textbf{Consistency} & \textbf{Fluency} & \textbf{Relevance} & \textbf{Average} \\
    \hline
    \multicolumn{6}{c}{\textit{Text matching-based metrics}}  \\
    \hline
    ROUGE-1 \cite{lin-2004-rouge} & 0.137 (8) & 0.111 (14) & 0.067 (13) & 0.228 (4) & 0.135 (9) \\
    ROUGE-2 \cite{lin-2004-rouge} & 0.110 (13) & 0.107 (15) & 0.054 (15) & 0.184 (13) & 0.113 (15) \\
    ROUGE-L \cite{lin-2004-rouge} & 0.109 (14) & 0.090 (16) & 0.067 (13) & 0.216 (8) & 0.120 (14) \\
    BLEU \cite{Papineni2002BleuAM} & 0.119 (11) & 0.126 (9) & 0.104 (8) & 0.185 (12) & 0.133 (10) \\
    METEOR \cite{lavie-agarwal-2007-meteor} & 0.112 (12) & 0.118 (12) & 0.079 (12) & 0.210 (10) & 0.129 (12) \\
    CHRF \cite{popovic-2015-chrf} & 0.168 (1) & 0.121 (10) & 0.086 (11) & 0.242 (3) & 0.154 (7) \\
    CIDEr \cite{Vedantam2015CIDErCI} & -0.003 (18) & 0.006 (18) & 0.037 (17) & -0.019 (18) & 0.005 (18) \\
    \hline
    \multicolumn{6}{c}{\textit{Embedding-based metrics}} \\
    \hline
    MoverScore \cite{zhao-etal-2019-moverscore} & 0.154 (4) & 0.134 (7) & 0.117 (4) & 0.224 (6) & 0.157 (6) \\
    SMS \cite{clark-etal-2019-sentence} & 0.144 (6) & 0.188 (2) & 0.133 (2) & 0.177 (14) & 0.160 (5) \\
    ROUGE-WE \cite{ng-abrecht-2015-better} & 0.087 (15) & 0.065 (17) & 0.012 (18) & 0.176 (15) & 0.085 (17) \\
    \hline
    \multicolumn{6}{c}{\textit{Model-based metrics}} \\
    \hline
    BERTScore \cite{Zhang*2020BERTScore:} & 0.126 (10) & 0.121 (10) & 0.113 (6) & 0.213 (9) & 0.143 (8) \\
    S$^3$ \cite{peyrard-etal-2017-learning} & 0.166 (2) & 0.113 (13) & 0.044 (15) & 0.227 (5) & 0.125 (13) \\
    BlANC \cite{Lita2005BLANCLE} & 0.084 (16) & 0.181 (4) & 0.099 (10) & 0.168 (16) & 0.133 (10) \\
    SUPERT \cite{gao2020supert} & 0.130 (9) & 0.259 (1) & 0.167 (1) & 0.204 (11) & 0.190 (1) \\
    SummaQA \cite{scialom-etal-2019-answers} & 0.062 (17) & 0.128 (8) & 0.101 (9) & 0.107 (17) & 0.099 (16) \\
    \hline
    \multicolumn{6}{c}{\textit{Proposed Metric}} \\
    \hline
    WIDAR$_1$ & 0.160 (3) & 0.178 (5) & 0.114 (5) & 0.254 (1) & 0.176 (2) \\
    WIDAR$_2$ & 0.138 (7) & 0.188 (2) & 0.108 (7) & 0.221 (7) & 0.163 (4) \\
    WIDAR$_L$ & 0.149 (5) & 0.176 (6) & 0.119 (3) & 0.250 (2) & 0.167 (3) \\
    \hline
    \end{tabular}%
    \vspace{-2em}
\end{table}

The fact that SUPERT \cite{gao2020supert} is a model-based metric that evaluates the quality of a summary by taking as input the generated summary and the input document might be the reason for it having high correlation scores with consistency and fluency. Since input document comprises of grammatically correct and factually rich sentences, high performances on fluency and consistency are to be expected. CHRF \cite{popovic-2015-chrf} and S$^3$ \cite{peyrard-etal-2017-learning} on the other-hand perform well in coherence and relevance; which can be somewhat credited to their evaluation strategy that computes information overlap between generated summary and reference summary. Since reference summary contains only the most significant information from the input document put together in a presentable manner, it results in high relevance and coherence scores. 
We believe that since WIDAR uses information overlap of generated summary with both the input document and the reference summary efficiently, it performs optimally in all the four characteristics. 

\vspace{-1em}
\subsection{Computational Time Analysis}
\vspace{-0.5em}
Table \ref{tab:Time analysis} shows the comparison of computational time taken by WIDAR with respect to 5 state-of-the-art models or embedding based metrics computed using a single reference summary. The experiment is conducted for 100 randomly chosen summaries for all the metrics\footnote{This experiment was conducted on a Tyrone machine with Intel's Xeon W‑2155 Processor having 196 Gb DDR4 RAM and 11 Gb Nvidia 1080Ti GPU. GPU was only used for BLANC, SUPERT, BERTScore and SummaQA evaluation metrics.}. 
It is noticed that WIDAR takes about 0.6$\%$ of the computational time as compared to the average time taken by all these 5 metrics, while giving similar performance. 

\vspace{-2.5em}
\begin{table}
  \centering
  \caption{Computation time taken by WIDAR and various model-based metrics.}
  \vspace{0.5em}
  \label{tab:Time analysis}
    \begin{tabular}{l@{\hskip 0.15in}|@{\hskip 0.15in}c}
    \hline
    \textbf{Metric} & \textbf{Time-taken} \\
    \hline
    BLANC \cite{Lita2005BLANCLE}  &1076.35 s \\
    SUPERT \cite{gao2020supert} &30.40 s \\
    MoverScore \cite{gao2020supert} &37.60 s\\
    BERTScore \cite{Zhang*2020BERTScore:} &1410.37 s\\
    SummaQA \cite{scialom-etal-2019-answers} &910.26 s\\
    \hline
    Average & 692.99 s \\
    \hline
    WIDAR$_L$ &3.96 s\\
 
    \hline
    \end{tabular}%
\end{table}
\vspace{-3em}

\vspace{-2.5em}
\begin{table}
  \centering
  \caption{Ablation Study.}
  \vspace{0.5em}
  \label{tab:ablation}
    \begin{tabular}{lc@{\hskip 0.1in}c@{\hskip 0.1in}c@{\hskip 0.1in}c}
    \hline
    \textbf{Metric}         & \textbf{Coherence} & \textbf{Consistency}
    & \textbf{Fluency} & \textbf{Relevance} \\
    \hline
    WIDAR$_{L} $ & 0.149 & 0.176 & 0.119 & 0.250 \\
    ROUGE-L$_W$ & 0.129 & 0.108 & 0.083 & 0.239 \\
    $-W_{red}$  & 0.105 & 0.075 & 0.064 & 0.201 \\
    $-W_{cov}$  & 0.119 & 0.115 & 0.087 & 0.218 \\
    ROUGE-L$_{SL} $ &0.102 & 0.087 & 0.062 & 0.204 \\
    \hline
    ROUGE-L  & 0.109 & 0.090 & 0.067 & 0.216 \\
    IDSS & 0.097 & 0.202 & 0.122 & 0.164 \\
    \hline
    \end{tabular}%
\end{table}
\vspace{-3em}

\subsection{Ablation Study}
\vspace{-0.5em}
WIDAR comprises of two key-components: (1) weighted ROUGE (ROUGE$_W$) between reference summary and generated summary and (2) similarity overlap (IDSS) between input document and generated summary. In order to establish the necessity of both of these components, we conduct an ablation study. When we consider only ROUGE-L$_W$, we notice a major drop in correlation with consistency ($38\%$) and fluency ($30\%$) (refer to the top two rows in Table \ref{tab:ablation}). We reason that consistency being the measure of factual correctness in the summary justifies the decrease in consistency scores. An argument can be made regarding fluency that since WIDAR is effectively a string-matching based technique; the input document usually comprises of sentences which are more grammatically sound than ones in reference summary \cite{fabbri2021summeval,kryscinski-etal-2019-neural}) could explain the drop in fluency scores. This argument can be further bolstered when comparing the correlation scores obtained for ROUGE-L and IDSS. IDSS uses ROUGE-L to compute information overlap between generated summary and input document, and ROUGE-L is used to compute information overlap between generated summary and reference summary. We can see that IDSS outperforms ROUGE-L in consistency (by $124\%$) and fluency (by $82\%$), supporting the previously mentioned argument. \par
If we remove $W_{red}$ from weighted ROUGE, we observe drops in coherence (by $18\%$) and relevance (by $15\%$) as expected; but we also observe that without these redundancy weights, correlation with consistency and fluency also drop by $30\%$ and $22\%$, respectively. Removing $W_{cov}$ however yields mixed results in an isolated setting. Yet, together with $W_{red}$, the weighted ROUGE is able to outperform the sentence-level baseline. This can be noticed from the relevance scores in Table \ref{tab:ablation}; ROUGE-L$_{SL}$ attains 0.204 score, while adding $W_{red}$ yields an increase to 0.218 (shown in row $-W_{cov}$) and adding $W_{cov}$ drops the score to 0.201 (shown in row $-W_{red}$). However, combining these two to obtain ROUGE-L$_W$ attains 0.239 score, better than in the case of the individual components.

\vspace{-1em}
\subsection{Study of Human Judgement Scores}
\vspace{-0.5em}
To analyze how humans have perceived these four characteristics of a summary, we compute and study the Kendall's Tau correlation coefficient between them. The results (refer to Table \ref{tab:humanscores}) revealed that coherence and relevance are moderately correlated, while other characteristic pairs do not yield any significant correlation score. This high correlation between coherence and relevance can be attributed to the fact that both relevance and coherence are related to non-redundancy. Coherence explicitly captures the non-redundancy in a summary, since a coherent summary must not have high information overlap across the sentences. Relevance on the other hand implicitly captures the notion of non-redundancy, since a summary that is highly relevant will cover up a major portion of input document, which is not achievable for a redundant summary. This reasoning can also be backed by the results from the ablation study (refer to Table \ref{tab:ablation}), where removing the redundancy weight ($W_{red}$) from weighted ROUGE affects both the relevance and the coherence scores, implying that humans directly or indirectly consider redundancy of sentences within summary while providing these scores.

\vspace{-2.5em}
\begin{table}
  \centering
  \caption{Kendall's Taus correlation between various summary characteristics.}
  \vspace{0.5em}
  \label{tab:humanscores}
    \begin{tabular}{lc@{\hskip 0.1in}c@{\hskip 0.1in}c@{\hskip 0.1in}c}
    \hline
    \textbf{}         & \textbf{Coherence} & \textbf{Consistency}
    & \textbf{Fluency} & \textbf{Relevance} \\
    \hline
    Coherence & 1.00 & 0.25 & 0.27 & 0.53 \\
    \hline
    Consistency & 0.25 & 1.00 & 0.38 & 0.27 \\
    \hline
    Fluency &0.27 & 0.38 & 1.00 & 0.23  \\
    \hline
    Relevance  &0.53 &0.27 & 0.23 & 1.00 \\
    \hline
    \end{tabular}%
\end{table}
\vspace{-3em}

\section{Conclusion} \label{sec:conclusion}
\vspace{-0.7em}
We propose a novel evaluation metric WIDAR that utilizes both input document and reference summary to estimate the quality of the generated summary. We discuss why metrics like ROUGE, METEOR, BLUE etc. that solely depend on reference summary for evaluation do not perform well in real-world situations. We illustrate how the proposed metric is able to outperform its predecessor, ROUGE, by a large margin, and is also able to achieve performance comparable to huge model-based metrics like BERTScore, S$^3$, SUPERT etc. We also perform an ablation study to establish the necessity of each component in the proposed metric. We believe that the community needs computationally fast and lightweight metrics like WIDAR that can work well in real-world situations. 

\textbf{Acknowledgement:}
Dr. Sriparna Saha gratefully acknowledges the Young Faculty Research Fellowship (YFRF) Award, supported by Visvesvaraya Ph.D. Scheme for Electronics and IT, Ministry of Electronics and Information Technology (MeitY), Government of India, being implemented by Digital India Corporation (formerly Media Lab Asia) for carrying out this research.


%
%
\bibliographystyle{splncs04}
\bibliography{main.bib}
\end{document}